\documentclass[]{spie}  %>>> use for US letter paper
\usepackage{amsmath,amsfonts,amssymb}
\usepackage{graphicx}
\usepackage[colorlinks=true, allcolors=blue]{hyperref}

\title{Dense Geometry Supervision for Underwater Depth Estimation}

\author[a]{Wenxiang Gu}
\author[a]{Lin Qi}
\affil[a]{School of Computer Science and Technology, Ocean University of China, Qingdao 266100}

\authorinfo{Further author information: (Send correspondence to Wenxiang Gu)\\Wenxiang Gu: E-mail: gwx1323@stu.ouc.edu.cn\\  Lin Qi: E-mail: qilin@ouc.edu.cn}

\begin{document} 
\maketitle

\begin{abstract}
The field of monocular depth estimation is continually evolving with the advent of numerous innovative models and extensions. However, research on monocular depth estimation methods specifically for underwater scenes remains limited, compounded by a scarcity of relevant data and methodological support. This paper proposes a novel approach to address the existing challenges in current monocular depth estimation methods for underwater environments. We construct an economically efficient dataset suitable for underwater scenarios by employing multi-view depth estimation to generate supervisory signals and corresponding enhanced underwater images. we introduces a texture-depth fusion module, designed according to the underwater optical imaging principles, which aims to effectively exploit and integrate depth information from texture cues. Experimental results on the FLSea dataset demonstrate that our approach significantly improves the accuracy and adaptability of models in underwater settings. This work offers a cost-effective solution for monocular underwater depth estimation and holds considerable promise for practical applications.
\end{abstract}

% Include a list of keywords after the abstract 

\section{INTRODUCTION}
\label{sec:intro}  % \label{} allows reference to this section

Monocular visual depth estimation finds significant applications across various domains such as underwater exploration\cite{Roznere_Li_2020}, marine research\cite{Raanan_Bellingham_Zhang_Kemp_Kieft_Singh_Girdhar_2018}, underwater engineering, and underwater vehicles\cite{Xanthidis_Karapetyan_Damron_Rahman_Johnson_O} (e.g., Autonomous Underwater Vehicles (AUVs) or Remotely Operated Vehicles (ROVs)). Although acoustic method is the dominant stream in underwater ranging, there has been a rising of optical methods with the advantages of high resolution and accuracy, such as underwater LiDAR \cite{Zhou_Li_Zhang_Liu_Zhou_Zhan_2021}, stereo and RGB-D cameras \cite{Lu_Zhang_Li_Zhou_Tadoh_Uemura_Kim_Serikawa_2017} and laser scanners\cite{sunqian}. However, these depth acquisition methods are complex and expensive in terms of devices or computational power required. In contrast, monocular depth estimation, due to its economic and efficiency, is a good alternative in underwater usages.

Monocular depth estimation techniques can be roughly divided into two research paths: supervised learning methods and unsupervised learning methods \cite{ming2021deep}\cite{lewei2024influence}. Under the supervised learning paradigm, existing monocular depth estimation methods exhibit significant limitations when dealing with underwater scenes \cite{Peng_Cosman_2017}, mainly due to the lack of sufficient high-quality annotated depth data as training resources \cite{Yang_Zhang_Wang_Xin_Hu_2022}. These limitations primarily manifest in regions with large depth values, where the model tends to confuse water medium with solid objects, leading to significant pixel-level depth estimation errors. To address the issue of scarce underwater environmental data, unsupervised method1s utilize frame-to-frame reprojection consistency as an intrinsic constraint to guide parameter optimization\cite{Godard_Aodha_Firman_Brostow_2019}. However, relying solely on neighboring frames for reprojection supervision struggles to address occlusion issues, making it difficult to effectively supervise regions with occlusions. Additionally, differences in imaging quality for the same object across adjacent frames, where an object in the scene may have poor imaging quality in one frame but good quality in the next, pose challenges for reprojection-based supervision.In underwater environments, this scenario is particularly prevalent due to the dynamic nature of these settings.

\begin{figure}[tb]
  \centering
  \includegraphics[width=\linewidth]{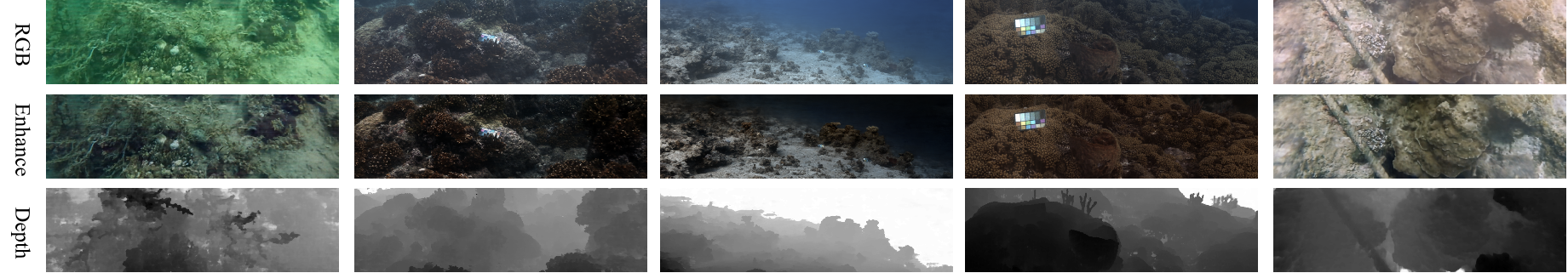}
  \caption{Here are some samples for generating the dataset.
  }
  \label{fig:example}
\end{figure}

To address the aforementioned issues, we innovatively propose a method inspired by unsupervised mechanisms. Our core objective is to mitigate existing problems of supervised methods in underwater scenes by generating more reliable monocular depth estimation supervision, specifically tailored for underwater environments. This approach enhances the applicability and performance of supervised learning methods in underwater scenes. Additionally, in underwater environments, although the medium affects imaging quality, textures in the captured images contain information about depth and water parameters\cite{Bekerman_Avidan_Treibitz_2020}. Therefore, effectively extracting this information from RGB images is crucial for monocular depth estimation in underwater scenes. We propose a fusion module that fully exploits and integrates texture and depth information based on underwater imaging principles. Through experiments conducted on the FLSea dataset\cite{Randall_Treibitz_2023}, we validate the effectiveness of this method. Experimental results demonstrate that our proposed method efficiently constructs depth supervision signals suitable for underwater conditions. Subsequently, we apply these automatically generated depth signals to fine-tune model parameters, significantly improving the accuracy and adaptability of the model in underwater scene recognition. Finally, we conduct ablation experiments for the texture-depth fusion module, further demonstrating the effectiveness of our module. Fig 1. illustrates the samples used for generating the dataset, which are derived from various scenarios. The first row of the figure presents images obtained from different perspectives through the Seethrough technology. The second row includes the enhanced versions of these images, which will serve as the input for the GBiNet model. The third row displays the estimated depth maps produced by the model.

Our main contributions can be summarized as follows:

\begin{itemize}
\item We address the issues of unsupervised methods by proposing the use of multi-view depth estimation methods to generate supervision signals, thereby constructing an underwater dataset suitable for training supervised methods.
\item We propose a texture-depth fusion module that fully exploits depth information embedded in texture information and achieves positive results on the FLSea dataset.

\end{itemize}

\section{METHODOLOGY}

\subsection{The Process of Constructing a Dataset\cite{Tosi_Tonioni_Gregorio_Poggi}}

 Initially, static scenes suitable for dataset creation are selected from numerous underwater videos. Subsequently, scene images are extracted from these videos, and SeathruNeRF\cite{Levy_Peleg_Pearl_Rosenbaum_Akkaynak_Korman_Treibitz} is employed for image enhancement and novel view synthesis. Finally, MVS(Multi-View Stereo) methods\cite{furukawa2015multi} are utilized to estimate depth information for the enhanced scene images.

\textbf{Image Collection and Preprocessing:} Both the new view synthesis technique based on NeRF\cite{mildenhall2021nerf}(Neural Radiance Fields) and the depth estimation method based on MVS are applicable to static scenes. During the actual scene image collection process, it is often necessary to filter out video frames of static scenes from the collected videos. For example, in underwater videos, dynamic elements such as swimming marine life and swaying seaweed often exist, so we should try to avoid these dynamic scenes as much as possible during image collection. Additionally, we should strive to cover scenes of different scales to make the constructed dataset more diverse and rich in information. After selecting suitable scene video segments, we need to appropriately sample the video frames to ensure a wide distribution of scene perspectives. At the same time, we remove frames with motion blur to ensure data quality. Finally, we retain 25 to 45 images as training data for Seathru-NeRF. In each scene, we use the COLMAP algorithm to estimate the camera's intrinsic matrix $K$ and the pose $E_i, i\in[1,M]$ (where $M$ is the number of images per scene) for each image. These data are indispensable elements in the training process of Seathru-NeRF.

\textbf{NeRF Rendering and Depth Generation:} We employed SeathruNeRF for weight training on a per-scene basis to generate dense viewpoint images along with their corresponding enhanced images. Each scene is supervised through the loss function described below; SeathruNeRF utilizes $s=\{s_i\}_{i=0}^{N}$ to represent a sampling sequence of a ray, where the $i$-th segment weight of an object is defined as:
\begin{equation}
\begin{gathered}
\omega_i^{obj}=T_i^{obj}\cdot(1-\exp(-\sigma_i^{obj}\delta_i)),
\label{eq:fuseRD}
\end{gathered}
\end{equation}
the weight sequence is denoted by $\pmb w =\{\omega_i^{obj}\}_{i=0}^N $, with the ground truth pixel values being $C^*$ \cite{Mildenhall_Hedman_Martin}. The complete loss function is given by:
\begin{equation}
\begin{gathered}
L=L_{recon}(\hat C,C^*)+L_{prop}(s,w)+\lambda L_{acc}(w),
\label{eq:fuseRD}
\end{gathered}
\end{equation}
where $\lambda = 0.0001$. This loss function effectively ensures the accuracy of synthesizing new viewpoints and the correctness of Seathru-NeRF's ray sampling point weights, thereby better distinguishing between medium and object. Subsequently, we utilize the scene-enhanced images obtained via Seathru-NeRF as input for GBiNet to acquire depth supervision information as accurately as possible, along with the corresponding confidence maps. 

\textbf{Post-processing:} Through the aforementioned steps, we obtained RGB images, enhanced images, and corresponding depth maps for multiple scenes. Based on the confidence maps generated by GBiNet\cite{mi2022generalized}, we filter out images with poor depth estimation and those exhibiting significant flaws. Simultaneously, we utilize the confidence maps to generate masks for the corresponding depth maps, ensuring that only reliable depth supervision information is retained during loss computation. Ultimately, the dataset comprises RGB images, corresponding enhanced images, depth supervision information, and masks for the depth maps across multiple scenes.

\begin{figure}[htbp]
\centerline{\includegraphics[width=\linewidth]{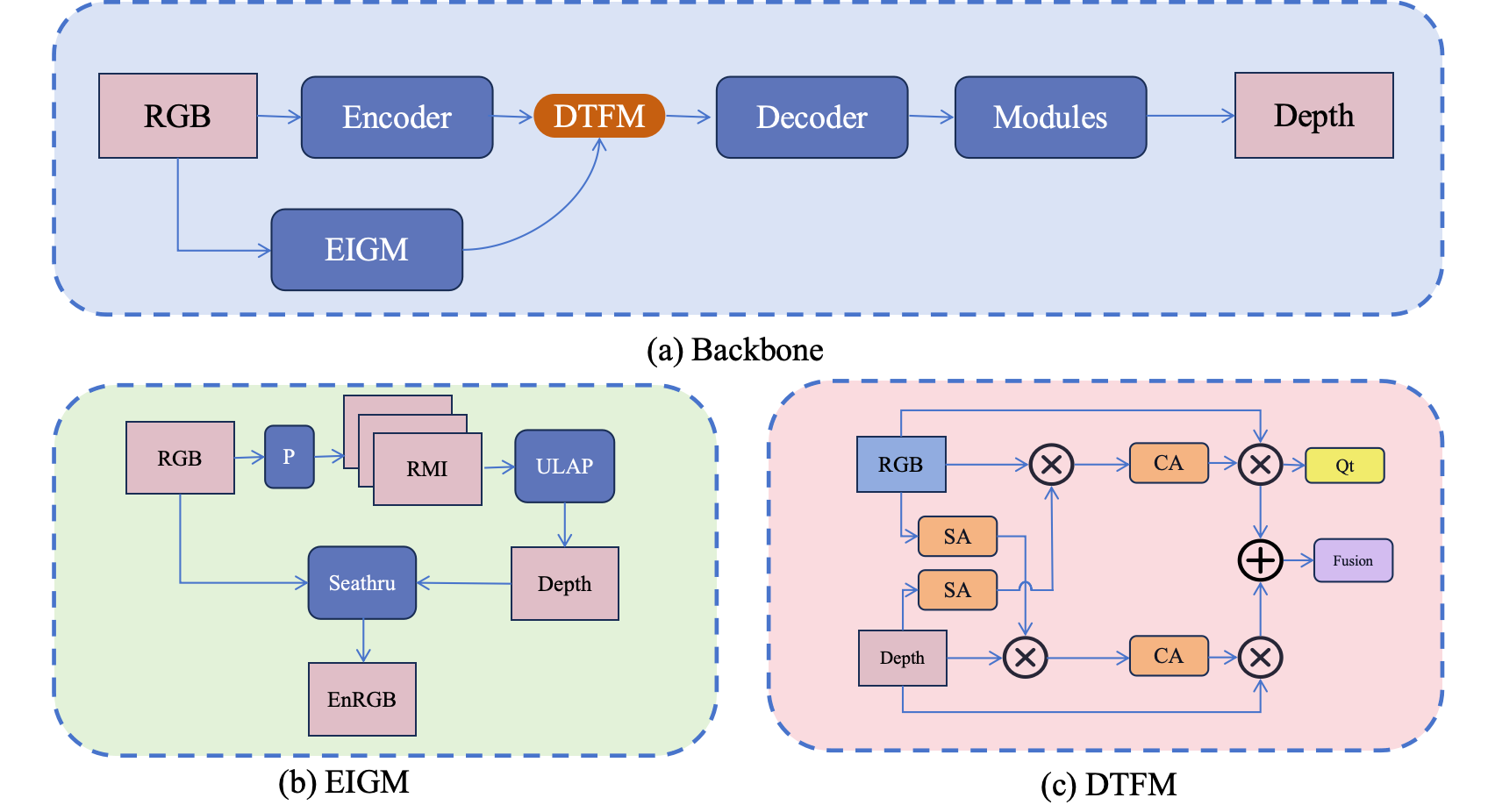}}
\caption{Integration of the Depth-Texture Fusion Module within the comprehensive mainstream architecture for monocular depth estimation.
}
\label{fig:example}
\end{figure}

\subsection{Texture-Depth Fusion Module} \label{BB}

To fully leverage the depth information inherent in underwater images, we propose the Underwater Texture-Depth Fusion Module, which achieves decoupling of underwater image enhancement and underwater depth estimation. Such an approach assists monocular depth estimation models designed for air medium in better adapting to underwater scenes.Fig 2. describes the use of the Depth-Texture Fusion Module within the monocular depth estimation backbone. Here, RGB denotes the monocular image requiring depth estimation, and Encoder refers to the feature extraction/encoder in different backbones. EIGM represents the Enhanced Image Feature Generation Module. As illustrated in section b of the figure, the RGB image undergoes processing (P) to convert from the RGB space to the RMI\cite{yu2023udepth} space, then ULAP(Underwater Light Attenuation Prio) generates the depth image, which is used by the Seathru algorithm to generate enhanced RGB texture features. DTFM stands for the Depth-Texture Fusion Module. The detailed implementation is shown in section c , where SA denotes Spatial Attention, and CA denotes Channel Attention. $Q_t$ represents the weight of the texture features derived from this module. Fusion represents the final combined features.

\textbf{Underwater Optical Imaging\cite{Akkaynak_Treibitz_2019}:} In underwater environments, the imaging process of an image can be described as follows:
\begin{equation}
\begin{gathered}
I_c = J_c e^{-\beta_c^D(V_D \cdot z)} + B_c^{\infty} (1 - e^{-\beta_c^B(V_B \cdot z)})
\label{eq:fuseRD}
\end{gathered}
\end{equation}

Here, $ z $ represents the distance along the line of sight between the camera and objects in the scene. $ B_c^{\infty} $ denotes the occluding light, $ J_c $ represents the unattenuated scene observed at the camera position if there is no attenuation along the $ z $ direction. Vectors $ V_D $ and $ V_B $ consist of variables related to water quality and depth. Sea-thru further suggests that $ \beta_c^D $ and $ \beta_c^B $ depend on wavelength, $ z $, and depth $ d $. Therefore, the unattenuated image $ J_c $ can be expressed as:
\begin{equation}
\begin{gathered}
J_c = (I_c - B_c) \cdot e^{\beta_c^D(\nu_D) \cdot z}
\label{eq:fuseRD}
\end{gathered}
\end{equation}

It was proposed in [2] that $ \beta_c^D $ is mainly influenced by the line-of-sight distance $ z $, while $ \beta^B_c $ is mainly influenced by the optical water quality type and illumination $ E $. As pointed out in [1], $ B_c^{\infty} $ grows exponentially with distance $ z $, reaching saturation when the scene reflectance $ \rho_c $ approaches 0 (all light is absorbed) or when illumination $ E $ approaches 0 (completely in shadow). Thus, the captured $ RGB $ intensity $ I_c $ approximates the backscattering intensity $ B_c^{\infty} $. For each of the 10 clusters uniformly divided by $ z $ (10 different depth segments), searching for the bottom 1% of $ RGB $ triplets in $ I_c $ yields these triplets, denoted as $ \Omega $. Therefore, $ \hat B(\Omega) \approx I_c(\Omega) $, and thus $ B_c $ for each cluster is estimated. Meanwhile, since $ \beta_c^B $ is mainly influenced by the optical water quality type and illumination $ E $, $ \beta_c^B $ can be estimated as:
\begin{equation}
\begin{gathered}
\hat \beta_c^D = -\log \hat E_c(z) / z
\label{eq:fuseRD}
\end{gathered}
\end{equation}
Consequently, estimating the direct attenuation coefficient $ \beta_c^D $ from image information becomes estimating the local illumination $ E_c(z) $. A specific estimation method for $ E_c(z) $ is provided in [2], which will not be reiterated here. Through the aforementioned steps, the necessary parameters for calculating the unattenuated image $ J_c $ are estimated.

\textbf{Underwater Light Attenuation Prior (ULAP)\cite{song2018rapid}:}  To extract more depth information from the enhancement process and decouple the depth estimation model from the image enhancement process, it is preferable not to rely on the roughly estimated depth from the monocular depth model backbone as a reference for Sea-thru's restored images. Therefore, we propose using ULAP to provide the necessary range map for Sea-thru. In various types of water bodies, the attenuation of red and near-infrared light underwater is significantly higher than that of shorter wavelengths in the visible spectrum. This phenomenon results in a rapid decrease in the intensity of the red channel with increasing distance in underwater scenes\cite{Pope_Fry_1997}. Based on this characteristic, we introduce the Underwater Light Attenuation Prior (ULAP). The computation of ULAP involves differential processing between the maximum intensity values of the blue and green channels (denoted as 
B and 
G, respectively) and the intensity value of the red channel (denoted as 
R), which is given by:
\begin{equation}
\begin{gathered}
d(x) = \mu_0 + \mu_1 \cdot \max(B(x),G(x)) + \mu_2 R(x)
\label{eq:fuseRD}
\end{gathered}
\end{equation}
where, $ \mu_0, \mu_1, \mu_2 $ are parameters learned through the loss function, and $ B(x), G(x), R(x) $ represent the values of the corresponding channels at pixel $ x $. $ d(x) $ is linearly correlated with the scene depth. By uniformly dividing $ I_c $ into 10 clusters using $ d(x) $ and estimating the local light source $ E_c(z) $ through $ d(x) $ and related pixel information, the enhanced image $ J_c $ can be computed using the method mentioned above.

\begin{figure*}[htbp]
\centerline{\includegraphics[width=\textwidth]{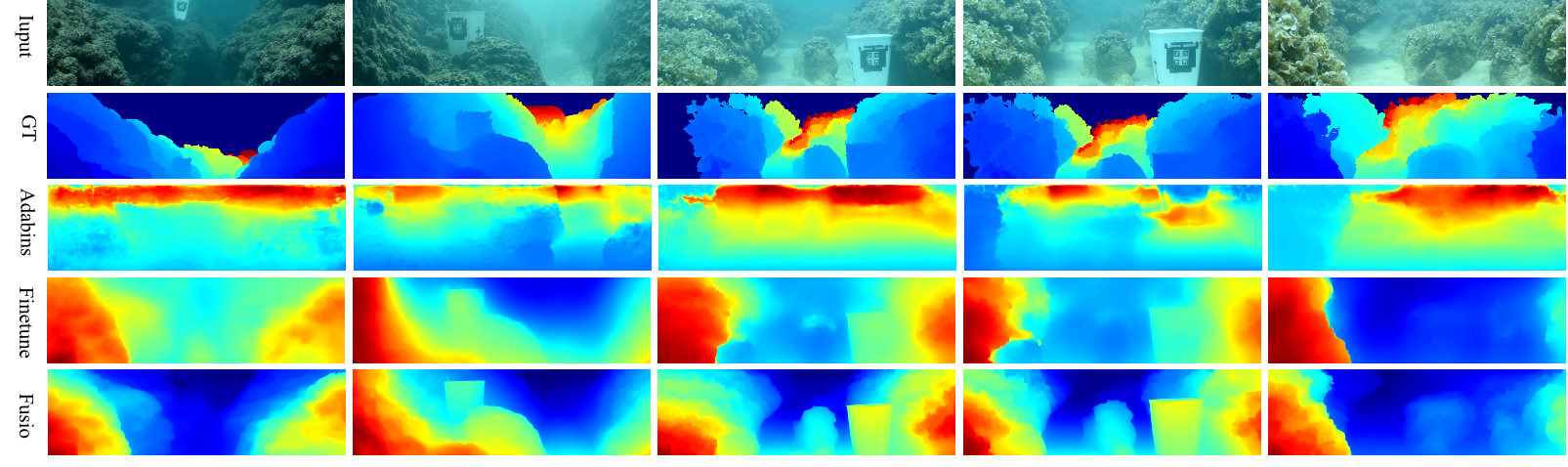}}
\caption{Depth Estimation Results for Different Schemes }
\label{fig}
\end{figure*}
\label{sec:typestyle}

\section{Implementation Details}\label{AA}
\textbf{Dataset Construction:} During the collection of underwater videos in static scenes, a total of 36 scenes were gathered, among which 8 scenes had invalid poses estimated by COLMAP, and 17 scenes had poor rendering quality using seathru-NeRF. Finally, 11 scenes with acceptable quality were obtained, with each scene uniformly rendered from 240 viewpoints. After removing some low-quality images, a total of 2131 data were obtained. The depth results were obtained by using GBiNet to predict the enhanced images. The output of GBiNet includes depth estimation corresponding to the images and confidence images corresponding to the depth maps. The rendering process of images and MVS depth estimation were both performed using an RTX 3090. When using SeathruNeRF for rendering, we followed the parameters suggested by the authors for scene image rendering. When predicting depth using GBiNet, the enhanced images and the poses estimated by COLMAP were used as input. The system does not predict point clouds but instead outputs depth images and their corresponding confidence maps.

\textbf{Fine-tuning Process:} In the actual fine-tuning process, we used multiple models to test the effectiveness of the dataset we constructed. At the same time, we also inserted the designed texture-depth fusion module into multiple networks for comparative experiments to verify the effectiveness of our module.

\textbf{Training Set:} We trained the initial weights using the KITTI dataset and then fine-tuned these weights using our custom-built dataset. Our constructed dataset comprises 11 scenes of varying sizes.

\textbf{Test Set:} We manually removed poor-quality data pairs (RGB-Depth) from FLSea, and then sampled uniformly from the remaining data pairs and from different scenes to obtain N pairs of data as our evaluation dataset. The standard metrics we used are displayed at the top of Table 1.\cite{Eigen_Puhrsch_Fergus_2014}.

\subsection{Results}\label{AA}
Table 1. presents a comparative analysis based on our constructed dataset and the Depth-Texture Fusion Module. The prefix before each model name indicates the dataset used:  "-Kitti" represents results obtained using only the Kitti dataset, "-UW" indicates results after fine-tuning with our constructed dataset, and "-F" denotes results obtained using both the Depth-Texture Fusion Module and the constructed dataset. The results reveal that model performance metrics improve significantly after fine-tuning with our dataset. Moreover, experiments incorporating the Depth-Texture Fusion Module demonstrate further enhancement in metrics, underscoring the utility of our dataset in facilitating model adaptation to underwater environments. These findings also highlight the effectiveness of the proposed Depth-Texture Fusion Module in improving single-image depth estimation in underwater scenes.

Fig 3. provides visual comparisons across different approaches. The first row shows the input RGB images, the second row displays the ground truth (GT), and the third row illustrates depth estimation results using Adabins. The fourth row presents the results obtained by Adabins after initial training on Kitti followed by fine-tuning with our constructed dataset. Finally, the last row exhibits the outcomes achieved using the Depth-Texture Fusion Module fine-tuned with the constructed dataset.

\begin{table}[htbp]
\caption{Quantitative results of the dataset and the Depth-Texture Fusion Module across different methods.}
\label{tab:performance}
\begin{center}
\begin{tabular}{|c|c|c|c|c|c|c|c|}
\hline
\rule[-1ex]{0pt}{3.5ex} \textbf{Model-Dataset} & \textbf{Abs Rel} & \textbf{Sq Rel} & \textbf{RMSE} & \textbf{RMSElog} & $\boldsymbol{\sigma} < \mathbf{1.25}$ & $\boldsymbol{\sigma} < \mathbf{1.25^2}$ & $\boldsymbol{\sigma} < \mathbf{1.25^3}$ \\
\hline
\rule[-1ex]{0pt}{3.5ex} NewCRFs\cite{yuan2022neural}-Kitti & 8.710 & 1.282 & 1.889 & 0.336 & 0.165 & 0.736 & 0.801 \\
\rule[-1ex]{0pt}{3.5ex} NewCRFs-UW & 1.251 & 0.764 & 1.382 & 0.372 & 0.662 & 0.915 & 0.963 \\
\rule[-1ex]{0pt}{3.5ex} NewCRFs-F & 0.473 & 0.233 & 0.702 & 0.357 & 0.715 & 0.923 & 0.972 \\
\hline
\rule[-1ex]{0pt}{3.5ex} IEBins\cite{shao2024iebins}-Kitti & 0.891 & 0.441 & 1.459 & 1.546 & 0.611 & 0.836 & 0.951 \\
\rule[-1ex]{0pt}{3.5ex} IEBins-UW & 0.215 & 0.177 & 0.617 & 0.367 & 0.725 & 0.883 & 0.957 \\
\rule[-1ex]{0pt}{3.5ex} IEBins-F & 0.102 & 0.157 & 0.483 & 0.187 & 0.796 & 0.963 & 0.982 \\
\hline
\rule[-1ex]{0pt}{3.5ex} AdaBins\cite{bhat2021adabins}-Kitti & 1.060 & 0.655 & 1.954 & 0.727 & 0.532 & 0.913 & 0.965 \\
\rule[-1ex]{0pt}{3.5ex} AdaBins-UW & 0.584 & 0.352 & 0.691 & 0.228 & 0.692 & 0.892 & 0.947 \\
\rule[-1ex]{0pt}{3.5ex} AdaBins-F & 0.357 & 0.243 & 0.435 & 0.152 & 0.727 & 0.913 & 0.953 \\
\hline
\rule[-1ex]{0pt}{3.5ex} Binsformer\cite{li2024binsformer}-Kitti & 3.587 & 1.719 & 1.773 & 0.527 & 0.362 & 0.932 & 0.973 \\
\rule[-1ex]{0pt}{3.5ex} Binsformer-UW & 1.091 & 0.939 & 1.092 & 0.302 & 0.573 & 0.917 & 0.965 \\
\rule[-1ex]{0pt}{3.5ex} Binsformer-F & 0.453 & 0.535 & 0.479 & 0.162 & 0.690 & 0.937 & 0.975 \\
\hline
\rule[-1ex]{0pt}{3.5ex} Va-depth\cite{liu2023va}-Kitti & 1.527 & 0.312 & 0.952 & 0.570 & 0.316 & 0.717 & 0.836 \\
\rule[-1ex]{0pt}{3.5ex} Va-depth-UW & 0.317 & 0.289 & 0.758 & 0.443 & 0.698 & 0.866 & 0.915 \\
\rule[-1ex]{0pt}{3.5ex} Va-depth-F & 0.244 & 0.240 & 0.413 & 0.292 & 0.721 & 0.935 & 0.974 \\
\hline
\end{tabular}
\end{center}
\end{table}

\section{Conclusion}\label{FAT}
We propose a method that combines multi-view depth estimation with underwater image enhancement to address the challenges in self-supervised underwater monocular depth estimation, such as inconsistent imaging quality and occlusion. Our approach also introduces a texture-depth fusion module, improving depth estimation accuracy by leveraging underwater optical imaging principles. Experimental results on our dataset demonstrate the effectiveness and significance of our framework in overcoming challenges in underwater depth estimation.

\newpage

% References
\bibliography{main} % bibliography data in report.bib
\bibliographystyle{spiebib} % makes bibtex use spiebib.bst

\end{document}